\documentclass[journal]{IEEEtran}

\usepackage{amsmath,amssymb,amsfonts}
\usepackage{mathrsfs}
\usepackage{algorithmic}
\usepackage{float}
\usepackage{graphicx}
\usepackage{textcomp}
\usepackage{xcolor}
\usepackage{stfloats}
\usepackage{subfigure}
\usepackage{array,mathtools}
\usepackage{multirow}
\usepackage{float}
\usepackage{bm}
\usepackage{algorithm}
\usepackage{booktabs}
\usepackage{cite}
\usepackage{verbatim}
\usepackage{color}

\usepackage[colorlinks,linkcolor=black,anchorcolor=black,citecolor=black,urlcolor=black]{hyperref}

\begin{document}

\title{Shape-Interpretable Visual Self-Modeling Enables Geometry-Aware Continuum Robot Control}

\author{Peng~Yu, Xin Wang, and Ning~Tan*

\thanks{* Corresponding author.}

\thanks{All the authors are with the School of Computer Science and Engineering, Sun Yat-sen University, Guangzhou 510006, China. (e-mail: yupeng6@mail2.sysu.edu.cn; wangx767@mail2.sysu.edu.cn; tann5@mail.sysu.edu.cn)}
}

\markboth{}%
{}
%


\maketitle

\begin{abstract}
Continuum robots possess high flexibility and redundancy, making them well suited for safe interaction in complex environments, yet their continuous deformation and nonlinear dynamics pose fundamental challenges to perception, modeling, and control. Existing vision-based control approaches often rely on end-to-end learning, achieving shape regulation without explicit awareness of robot geometry or its interaction with the environment.
Here, we introduce a shape-interpretable visual self-modeling framework for continuum robots that enables geometry-aware control. Robot shapes are encoded from multi-view planar images using a Bézier-curve representation, transforming visual observations into a compact and physically meaningful shape space that uniquely characterizes the robot's three-dimensional configuration. Based on this representation, neural ordinary differential equations are employed to self-model both shape and end-effector dynamics directly from data, enabling hybrid shape-position control without analytical models or dense body markers.
The explicit geometric structure of the learned shape space allows the robot to reason about its body and surroundings, supporting environment-aware behaviors such as obstacle avoidance and self-motion while maintaining end-effector objectives. Experiments on a cable-driven continuum robot demonstrate accurate shape-position regulation and tracking, with shape errors within 1.56\% of image resolution and end-effector errors within 2\% of robot length, as well as robust performance in constrained environments.
By elevating visual shape representations from two-dimensional observations to an interpretable three-dimensional self-model, this work establishes a principled alternative to vision-based end-to-end control and advances autonomous, geometry-aware manipulation for continuum robots.

\end{abstract}


\IEEEpeerreviewmaketitle

\section{Introduction}

Continuum robots represent a distinct class of robotic systems whose motion is generated not through discrete rigid joints but via continuous deformation of flexible backbones. This morphological paradigm enables motions reminiscent of biological appendages such as octopus tentacles~\cite{Xie2023SciRob}, snakes~\cite{Qin2022SoRo}, and elephant trunks~\cite{Ma2023RAL}, endowing continuum robots with high kinematic redundancy, intrinsic compliance, and remarkable adaptability to unstructured environments. These properties make them particularly attractive for tasks that demand safe, smooth, and dexterous interaction within confined, complex, or uncertain spaces, including minimally invasive surgery~\cite{Wang2023TMECH}, disaster response~\cite{Wen2023JMR}, inspection of complex pipelines and industrial equipment~\cite{Yang2024TRO}, and human-robot collaboration~\cite{Chen2025RAL}. Yet, the very characteristics that grant continuum robots their versatility, i.e., continuous deformation, structural flexibility, and high redundancy, also introduce pronounced nonlinearity, strong coupling, and significant parametric uncertainty. As a result, achieving reliable perception, modeling, and control of continuum robots remains a fundamental and open challenge.

Over the past decades, extensive research has been devoted to the end-effector control of continuum robots. Both model-based and data-driven strategies have been explored, including inverse kinematics solvers under constant-curvature assumptions~\cite{Qiu2025IJRR}, finite-element-based dynamic models~\cite{Li2025TRO}, and Cosserat-rod-based kinetostatic formulations~\cite{Lilge2023TRO}. While these approaches have demonstrated impressive performance, their effectiveness is often limited by the strong nonlinearity, high coupling, and parametric uncertainty inherent to continuum robots. To mitigate modeling inaccuracies, a growing body of work has shifted toward model-free and data-driven control~\cite{Thuruthel2017SoRo}. Early studies exploited online Jacobian estimation~\cite{Yip2014TRO,Tan2025IJRR}, while more recent efforts leveraged Koopman operator~\cite{Bruder2021TRO,Thamo2023RAL,Wang2023RAL,Haggerty2023SR,Yu2025IJRR} and neural ordinary differential equation (NODE)~\cite{Kasaei2023ICRA,Kasaei2025ICRA} to learn robot dynamics directly from data. These learning-based approaches significantly simplify controller design and improve data efficiency, enabling accurate end-effector control even in the presence of substantial uncertainties~\cite{Yu2026TRO}.

However, end-effector control alone is insufficient for many real-world tasks involving continuum robots. Due to their extended and deformable bodies, safe and effective operation in cluttered or confined environments often requires the robot to actively regulate its body shape~\cite{Almanzor2023TRO,dupont2022IEEE,yeshmukhametov2022AS}, rather than merely positioning its tip. Consequently, shape control has emerged as a critical capability complementary to end-effector control.

\begin{figure*}[tbp]
    \centering
    \includegraphics[width=0.9\linewidth]{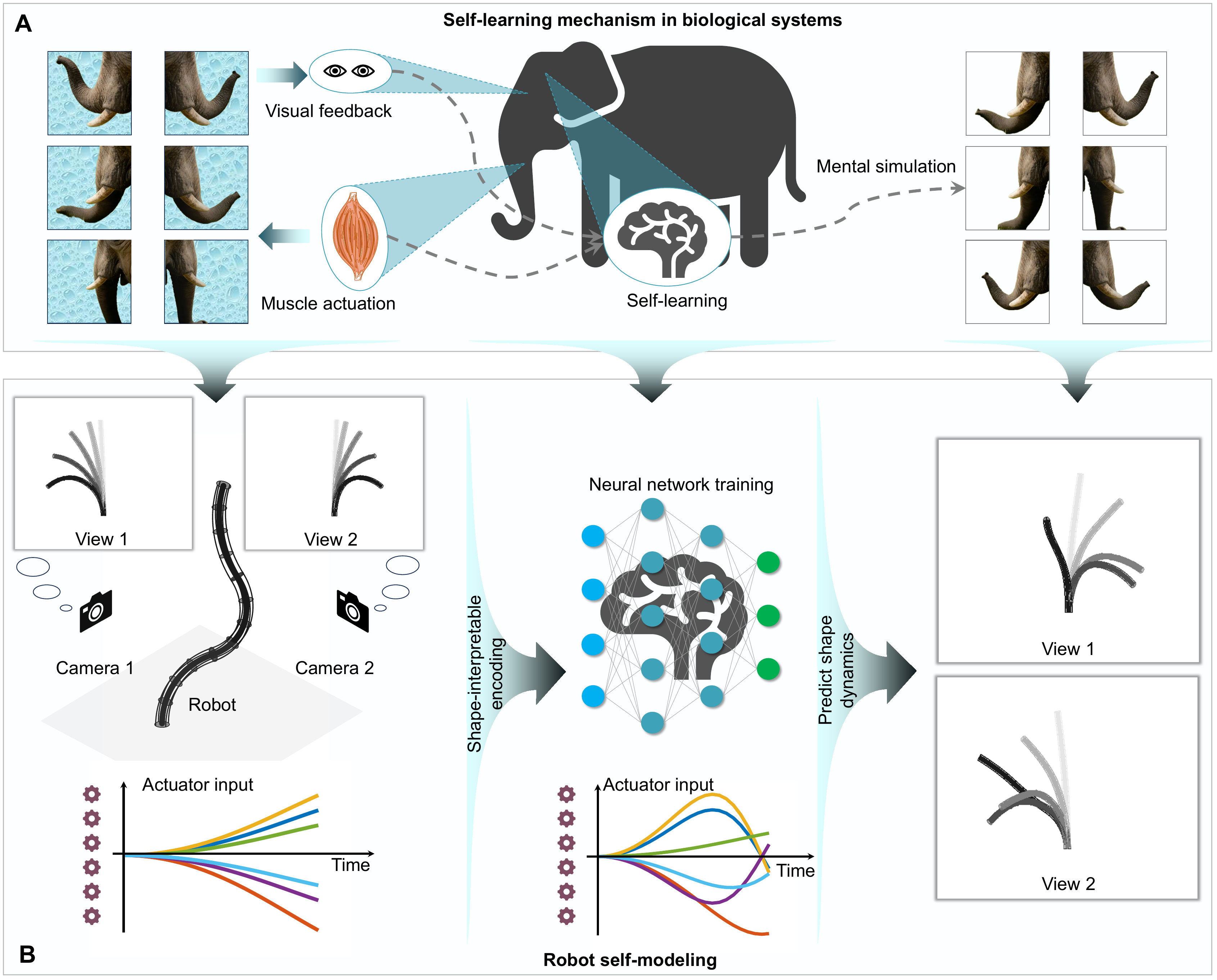}
    \caption{Conceptual illustration of the bio-inspired self-modeling mechanism. (A) Self-learning mechanism in biological systems. An elephant controls the motion of its trunk through muscular actuation and acquires corresponding trunk shape feedback via binocular vision, thereby learning the relationship between muscle activation and trunk morphology. Based on this learned internal model, the elephant can anticipate the shape changes induced by its own motor commands and accordingly control the trunk to perform complex manipulation tasks. (B) The continuum robot is actuated using arbitrary temporal input signals, while image data of the robot are recorded at each time instant from two different views. The image data are parameterized and encoded together with the actuation signals to form a dataset for neural network training. After training, the continuum robot can exploit the learned network to autonomously predict the visual shape dynamics from different views under given input temporal signals.}
    \label{fig:shape learning}
\end{figure*}

Existing shape control approaches for continuum robots can be broadly classified into three categories according to their shape representation paradigms: model-based methods, data-driven methods based on discrete shape points, and vision-based end-to-end learning methods. 
Model-based methods typically rely on explicit physical and geometric models, where shape is represented through analytical constructs, such as piecewise constant curvature models~\cite{Long2025IROS}, beam theories~\cite{Mbakop2021RAL,Mbakop2024TRO}, or Cosserat rod formulations. These models have been successfully integrated with optimization or predictive control frameworks to regulate robot shape and avoid obstacles. Nevertheless, in practice, accurately modeling continuum robots remains extremely challenging due to material nonlinearities, friction, hysteresis, and unmodeled external interactions. As a result, purely model-driven approaches often suffer from limited robustness and degraded performance in real-world scenarios.
To alleviate modeling dependence, data-driven shape control methods based on discrete feature points have been proposed~\cite{Gandhi2023IROS,Ouyang2018TMECH,Lai2020AIM,Singh2023Acc,Junfeng2023IJACSA,Shen2024RAL,Cao2025TRO,Chen2025TRO}. These methods represent the robot shape using a set of labeled points distributed along the backbone and regulate their positions jointly. While effective, such approaches typically require dense physical markers~\cite{Xu2023TMECH}, careful camera calibration, or expensive depth sensors. These requirements increase system complexity, interfere with the robot's intrinsic compliance, and limit practical deployment.
More recently, vision-based end-to-end learning methods have attracted increasing attention. By directly mapping images to control commands using deep neural networks~\cite{Almanzor2023TRO,MarquesMonteiro2024FRAI,Tang2026SA}, these approaches eliminate the need for explicit shape modeling or marker placement. Although they exhibit strong robustness to visual disturbances, most existing methods are restricted to planar shape control and suffer from fundamental limitations: (\romannumeral1) The learned shape representations are implicit, providing no explicit geometric awareness of the robot body or its distance to the environment; (\romannumeral2) single-view representations suffer from non-unique mappings, where different three-dimensional shapes can produce identical two-dimensional projections. As a consequence, such methods struggle to support flexible operations such as obstacle avoidance or self-motion through shape regulation.

Despite their differences, existing shape control paradigms share a common limitation: they lack a principled, interpretable, and learnable representation of the robot's full three-dimensional shape. In previous works, shape is either over-constrained by inaccurate models, under-represented by sparse measurements, or implicitly encoded in black-box models, hindering generalization, interpretability, and flexible reuse across tasks.
This observation motivates a fundamentally different perspective: rather than prescribing or bypassing shape models, can a continuum robot learn its own three-dimensional shape dynamics directly from multi-view observations, and exploit it for control?

\begin{figure*}[tbp]
    \centering
    \includegraphics[width=0.95\linewidth]{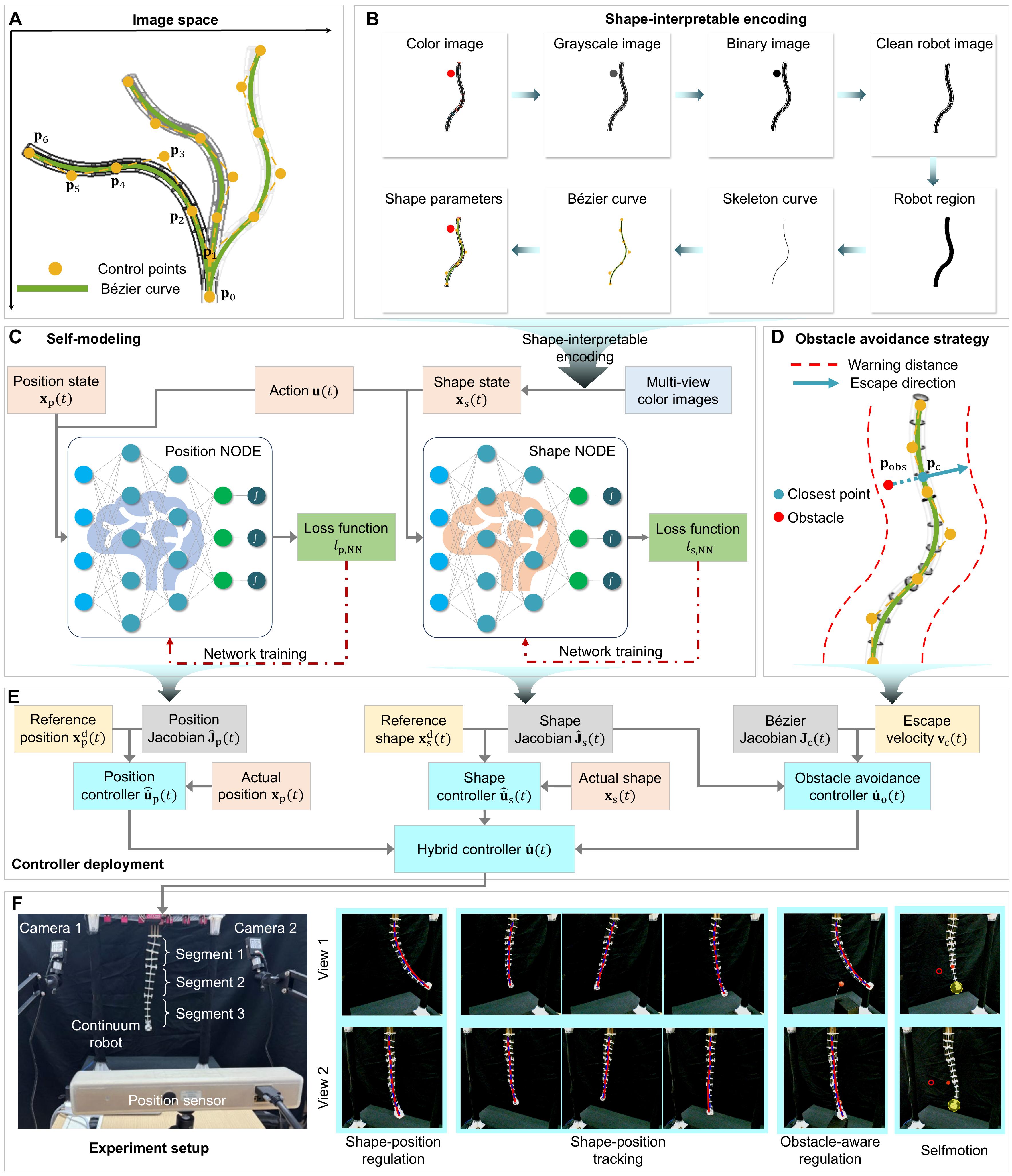}
    \caption{Shape-interpretable visual self-modeling and geometry-aware control framework for continuum robots. (A) The robot shape is represented using a piecewise Bézier curve, which captures global shape characteristics with a small number of control points. (B) The robot region is first extracted using image processing techniques, and the skeleton curve is obtained via morphological operations. The skeleton is then fitted by Bézier curves, and the resulting control points are used as shape parameters, enabling an interpretable shape encoding. (C) Position NODE and shape NODE models are trained using the collected shape, position, and actuation data to achieve shape-position self-modeling. (D) When an environmental obstacle enters a predefined warning distance, the obstacle avoidance mechanism is activated. A repulsive escape velocity is applied at the point on the robot body closest to the obstacle, and this local escape velocity is mapped to a global shape variation, thereby realizing obstacle avoidance through shape control. (E) The trained position NODE and shape NODE are used to estimate the position Jacobian and shape Jacobian, respectively. A hybrid controller is then constructed by integrating the position controller, shape controller, and obstacle avoidance controller. (F) The proposed control framework is validated on a three-segment continuum robot and is capable of performing shape-position regulation, shape-position tracking, obstacle-aware regulation, and self-motion tasks.}
    \label{fig:method}
\end{figure*}

Biological systems provide compelling evidence that explicit analytical models are not a prerequisite for controlling highly deformable bodies. For instance, an elephant can dexterously manipulate objects, avoid obstacles, and reconfigure its trunk despite its extreme redundancy and compliance. Such capabilities emerge through experience-driven self-learning, whereby the animal acquires an internal representation of the relationship between actuation and body deformation through efferent action signals and afferent visual feedback~\cite{MarquesMonteiro2024FRAI}. Inspired by this self-modeling paradigm, as shown in Fig.~\ref{fig:shape learning}, this work investigates a vision-based self-modeling approach for continuum robot shape control under minimal sensing assumptions. Instead of relying on analytical models, physical markers, or camera calibration, the proposed method enables a robot to autonomously learn its own shape dynamics purely from multi-view visual observations. While each single view provides only a two-dimensional representation, the combination of multiple distinct views establishes a unique correspondence to the robot's three-dimensional shape without explicit 3D reconstruction. The detailed self-modeling and control framework are shown in Fig.~\ref{fig:method}. The robot backbone in each view is parameterized using Bézier curves, yielding a compact, geometry-aware, and interpretable shape representation. Based on the representation, an NODE model is employed to learn the shape dynamics in visual spaces, forming an implicit self-model of the robot's three-dimensional shape evolution. In parallel, a separate NODE-based model is learned for end-effector position dynamics. Building upon the learned models, a unified Jacobian-based shape-position hybrid control framework is developed. Within this framework, shape regulation naturally enables flexible behaviors such as obstacle avoidance and self-motion, while maintaining accurate end-effector positioning.

Compared with existing approaches, the proposed method advances continuum robot control along several critical dimensions. (\romannumeral1) Unlike model-based strategies that rely on accurate analytical descriptions, the proposed framework eliminates the need for explicit physical modeling through vision-based self-modeling. (\romannumeral2) In contrast to data-driven shape control methods that depend on sparse or densely labeled body points, the proposed approach achieves marker-free shape representation using only low-cost monocular visual observations. (\romannumeral3) Unlike vision-based end-to-end control methods that implicitly encode shape in opaque latent spaces and are typically limited to planar control, the proposed framework explicitly represents robot shape in a geometrically meaningful form and naturally extends shape regulation from two-dimensional visual projections to consistent three-dimensional shape control. 

The main contributions of this work are threefold. (\romannumeral1) We propose a shape-interpretable visual self-modeling framework that enables a continuum robot to autonomously learn its own shape dynamics directly from multi-view visual observations, without requiring analytical models, physical markers, or camera calibration. (\romannumeral2) We develop a geometry-aware shape-position control strategy that explicitly exploits the learned shape representation to regulate body deformation, supporting flexible behaviors such as obstacle avoidance and self-motion. (\romannumeral3) We validate the effectiveness of the proposed approach through comprehensive experiments on a cable-driven continuum robot, demonstrating accurate shape-position regulation, reliable environment interaction, and clear performance advantages over a representative vision-based end-to-end control method.

\section{Preliminary}

This section presents essential background knowledge on the shape-position dynamics of continuum robots and Bézier curve for shape characterization.

\subsection{Shape-Position Dynamics}

For a continuum robot with $m$ actuators, its shape dynamics can be formulated as
\begin{equation}
\dot{\mathbf{x}}_\mathrm{s}(t) = f_\mathrm{s} (\mathbf{x}_\mathrm{s}(t), \mathbf{u}(t))
\label{eq:shape dynamics}
\end{equation}
where $\mathbf{x}_\mathrm{s}(t) \in \mathbb{R}^k$ denotes a $k$-dimensional shape state vector of the robot at time $t$, which is obtained by encoding robot images from different views, $\dot{\mathbf{x}}_\mathrm{s}(t)$ represents the first-order time derivative of $\mathbf{x}_\mathrm{s}(t)$, $\mathbf{u}(t)\in\mathbb{R}^m$ is the actuation input vector, and $f_\mathrm{s}$ is a nonlinear function that characterizes the shape dynamics.

Likewise, its end-effector dynamics is typically formulated as
\begin{equation}
\dot{\mathbf{x}}_\mathrm{p}(t) = f_\mathrm{p} (\mathbf{x}_\mathrm{p}(t), \mathbf{u}(t))
\label{eq:position dynamics}
\end{equation}
where $\mathbf{x}_\mathrm{p}(t) \in \mathbb{R}^n$ denotes the end-effector state (which corresponds to the end-effector position in this work) in the $n$-dimensional task space, $\dot{\mathbf{x}}_\mathrm{p}(t)$ represents the first-order time derivative of $\mathbf{x}_\mathrm{p}(t)$, and $f_\mathrm{p}$ is a nonlinear function that characterizes the end-effector dynamics.

The objective of this work is to identify an appropriate method for encoding the three-dimensional shape of the robot to facilitate visual shape-dynamics self-modeling, and to learn the aforementioned unknown shape dynamics~\eqref{eq:shape dynamics} and end-effector dynamics~\eqref{eq:position dynamics}, thereby enabling the design of a hybrid shape-position control strategy.

\subsection{Bézier Curve}

This paper attempts to determine a unique three-dimensional configuration of continuum robots using their multi-view planar images. The planar shape of a continuum robot is typically represented as a piecewise curve. To describe the shape of the continuum robot in the plane, Bézier curves are adopted in this work due to their explicit geometric interpretation and simple polynomial form. As illustrated in Fig.~\ref{fig:method}, the curve shape can be directly controlled by only a few control points~\cite{Xu2023TMECH}, which facilitates intuitive modeling and efficient enforcement of continuity constraints between adjacent segments, making Bézier curves particularly suitable for compact and interpretable modeling of planar multi-segment curves.

Consider a planar curve to be represented by $M$ quadratic Bézier segments, each defined on a local parameter $s\in[0,1]$. A single quadratic Bézier curve is defined by three control points $\mathbf{p}_0, \mathbf{p}_1, \mathbf{p}_2 \in \mathbb{R}^2$, and is parameterized as
\begin{equation}
\mathbf b(s) = (1-s)^2 \mathbf{p}_0 + 2(1-s)s \mathbf{p}_1 + s^2 \mathbf{p}_2
\end{equation}
where $\mathbf b(s)$ describes the planar coordinates of an arbitrary point on the curve with $s=0$ and $s=1$ corresponding to the strat point $\mathbf{p}_0$ and the end point $\mathbf{p}_2$, respectively. Accordingly, for an $M$-segment Bézier curve, the $i$-th segment is defined by
\begin{equation}
\mathbf b_i(s) = (1-s)^2 \mathbf{p}_{2i} + 2(1-s)s \mathbf{p}_{2i+1} + s^2 \mathbf{p}_{2i+2}
\label{eq:multi-segment bezier}
\end{equation}
where $i=0,\cdots,M-1$.

\section{Method}\label{method}

This section elaborates the proposed method for visual shape-dynamics self-modeling and hybrid shape-position control of continuum robots.

\subsection{Shape Encoding}

In this work, the complex planar configuration of the robot is parameterized using a small number of control points. Therefore, the first challenge is how to encode a two-dimensional image into control-point coordinates. As illustrated in Fig.~\ref{fig:method}, the color image is first converted into a grayscale image and then binarized. For simplicity, we assume that the binarization results for the robot region and the background are opposite, and that the robot region corresponds to the largest connected foreground component. Under this assumption, the image can be filtered to retain only the robot region. Then, a morphological closing operation is applied to obtain a clean robot region. Alternatively, more advanced semantic segmentation algorithms could be employed to directly identify the robot region and remove this assumption, which is, however, not the focus of this work. Next, a thinning operation is performed to extract the skeleton curve of the robot region~\cite{Lam1992TPAMI}. These steps can be easily implemented using the relevant functions available in MATLAB.

Subsequently, the skeleton curve is fitted using a Bézier curve via a least-squares approach, from which the control-point coordinates are obtained. The skeleton can be represented by an ordered set of points $\mathcal{Q} = \{\mathbf{q}_j \in \mathbb{R}^2 | j=1,\cdots,N\}$, where $\mathbf{q}_j$ denotes the coordinate of $j$-th pixel representing the skeleton in the image space. We take a three-segment Bézier curve as an example to illustrate how to fit the robot skeleton curve. This procedure can be easily extended to cases with fewer or more segments. According to the definition of quadratic Bézier curve~\eqref{eq:multi-segment bezier}, a three-segment quadratic Bézier curve is fully determined by seven control points $\{\mathbf{p}_j \in \mathbb{R}^2| j = 0,\cdots,6\}$. The point set $\mathcal{Q}$ is uniformly divided into three ordered subsets $\mathcal{Q}=\mathcal{Q}_1 \cup \mathcal{Q}_2 \cup \mathcal{Q}_3$, each corresponding to one Bézier segment. The endpoint control points are directly determined from the data:
\begin{equation}
\mathbf{p}_0 = \mathcal{Q}_1^{(1)}, \mathbf{p}_2 = \mathcal{Q}_1^{(\mathrm{end})}, \mathbf{p}_4 = \mathcal{Q}_2^{(\mathrm{end})}, \mathbf{p}_6 = \mathcal{Q}_3^{(\mathrm{end})}
\end{equation}
where $\mathcal{Q}_1^{(1)}$ denotes the first point of the first subset, $\mathcal{Q}_1^{(\mathrm{end})}$, $\mathcal{Q}_2^{(\mathrm{end})}$ and $\mathcal{Q}_3^{(\mathrm{end})}$ denote the last points of the three subsets, respectively. For each quadratic Bézier segment, only the intermediate control point is unknown. Consider the generic segment defined by~\eqref{eq:multi-segment bezier}, where $\mathbf{p}_{2i}$ and $\mathbf{p}_{2i+2}$ are known endpoints, and $\mathbf{p}_{2i+1}$ is the intermediate control point to be estimated. Given the corresponding skeleton points $\mathcal{Q}_i = \{\mathbf{q}_j^i\}_{j = 1}^{n_i}$ of the $i$-th segment, where $\mathbf{q}_j^i$ is the $j$-th skeleton point in the $i$-th segment and $n_i$ is the number of skeleton points in the segment, a normalized parameter is assigned as
\begin{equation}
s_j = \frac{j-1}{n_i - 1}, j = 1, \cdots, n_i.
\end{equation}
Rearranging the Bézier equation yields a linear relation with respect to $\mathbf{p}_{2i+1}$:
\begin{equation}
2(1-s_j) s_j \mathbf{p}_{2i+1} = \underbrace{\mathbf{q}_j^i - (1-s_j)^2 \mathbf{p}_{2i} - s_j^2 \mathbf{p}_{2i+2}}_{\mathbf{r}_j}.
\end{equation}
Stacking all skeleton points of the $i$-th segment leads to a linear least-squares problem:
\begin{equation}
\min_{\mathbf{p}_{2i+1}} \sum_{j=1}^{n_i} \|2(1-s_j)s_j \mathbf{p}_{2i+1} - \mathbf{r}_j\|^2 .
\end{equation}
The optimal intermediate control point $\mathbf{p}_{2i+1}$ is obtained by solving this linear least-squares problem. Applying this procedure independently to the three segments yields the control points $\mathbf{p}_{1}$, $\mathbf{p}_{3}$ and $\mathbf{p}_{5}$.

After fitting the robot shape using Bézier curves and obtaining the control-point coordinates, the next issue is how to define the shape parameter vector $\mathbf{x}_\mathrm{s}$. As shown in Fig.~\ref{fig:method}, the starting point $\mathbf{p}_{0}$ corresponding to the robot base is fixed. therefore, only the remaining control points need to be considered as dynamic variables. Accordingly, the planar shape feature matrix of a robot image is defined as the concatenation of the coordinate vectors of the remaining control points:
\begin{equation}
\mathbf{S} = [\mathbf{p}_{1},\cdots,\mathbf{p}_{2M}] \in \mathbb{R}^{2\times 2M}.
\end{equation}
Since the three-dimensional shape of the robot is uniquely determined in this work using planar images from $K$ different views, the robot's three-dimensional shape feature can be represented by 
\begin{equation}
\bar{\mathbf{S}} = [\mathbf{S}_1,\cdots,\mathbf{S}_K]^{\mathsf{T}} \in \mathbb{R}^{2MK \times 2}
\end{equation}
where $\mathbf{S}_1$ and $\mathbf{S}_K$ denote the planar shape feature matrices of the robot obtained from view 1 and view $K$, respectively. Finally, the three-dimensional shape feature vector of the robot can be expressed as 
\begin{equation}
\mathbf{x}_\mathrm{s} = \mathrm{vec}(\bar{\mathbf{S}}) \in \mathbb{R}^{4MK}
\end{equation}
where $\mathrm{vec}(\cdot)$ denotes the vectorization operation that stacks the columns of a matrix into a single vector. Since a minimum of two views is sufficient to uniquely determine the robot's three-dimensional shape, we set $K=2$ in this study. In practical applications, additional views can be incorporated to address occlusions or limited visibility in certain views, thereby enhancing the robustness of the proposed method.

Through the above procedure, robot images captured from different views can be encoded into a shape feature vector, which facilitates subsequent learning and control.

\subsection{Data-Driven Self-Modeling}

We next introduce how the robot's shape dynamics and position dynamics can be learned in a data-driven manner. This work primarily adopts neural ordinary differential equations (NODEs) to achieve self-modeling. Existing studies have demonstrated that this framework is more data-efficient~\cite{Kasaei2023ICRA,Yu2026TRO}, requiring much fewer training samples, than traditional neural network-based approaches~\cite{Thuruthel2017SoRo} or the more recently developed Koopman-operator-based methods~\cite{Bruder2021TRO}, which is particularly advantageous for learning high-dimensional shape features.

To model the shape dynamics~\eqref{eq:shape dynamics}, a time-dependent neural network (NN) is utilized to solve the following initial value problem:
\begin{equation}
    \left\{
        \begin{aligned}
    \frac{\partial \mathbf{x}_\mathrm{s}(t)}{\partial t} &= f_{s,\mathrm{NN}}(\mathbf{x}_\mathrm{s}(t), \mathbf{u}(t), t), \  t \ge t_0 \\ 
    \mathbf{x}_\mathrm{s}(t_0) &= \mathbf{x}_\mathrm{s,0} \\ 
    \mathbf{u}(t_0) &= \mathbf{u}_\mathrm{0} 
        \end{aligned}
    \right.
\end{equation}
where $f_{\mathrm{s},\mathrm{NN}}$ is the NN-based approximation of $f_\mathrm{s}$, $\mathbf{x}_\mathrm{s,0}$ and $\mathbf{u}_\mathrm{0}$ are initial values of $\mathbf{x}_\mathrm{s}(t)$ and $\mathbf{u}(t)$ at the initial time $t_0$, respectively. Given that we have the knowledge of $f_{\mathrm{s},\mathrm{NN}}$ and initial values, the estimated solution to the unknown shape dynamics~\eqref{eq:shape dynamics} at any time instant $t^+ \ge t_0$ is computed as
\begin{equation}
\hat{\mathbf{x}}_\mathrm{s}(t^+) = \mathbf{x}_\mathrm{s}(t_0) + \int_{t_0}^{t^+} f_{\mathrm{s},\mathrm{NN}}(\mathbf{x}_\mathrm{s}(t), \mathbf{u}(t), t) \mathrm{d} t
\end{equation}
which can be computed using standard numerical ordinary differential equation solvers (ODES) like the Runge-Kutta or Adams-Bashforth families of algorithms~\cite{Kasaei2023ICRA}. Therefore, the estimated solution is also denoted as 
\begin{equation}
\hat{\mathbf{x}}_\mathrm{s}(t^+) = \mathrm{ODES}\left(f_{\mathrm{s},\mathrm{NN}}, \mathbf{x}_\mathrm{s}(t_0), \mathbf{u}(t_0),[t_0,t^+]\right).
\end{equation}
To train the neural network $f_{\mathrm{s},\mathrm{NN}}$, the loss function is defined as
\begin{equation}
l_{\mathrm{s},\mathrm{NN}} = \|\hat{\mathbf{x}}_\mathrm{s}(t^+) - \mathbf{x}_\mathrm{s}(t^+)\|.
\end{equation}
To perform the training, we need to collect data pairs of the form $\{\mathbf{x}_\mathrm{s}(t_0), \mathbf{u}(t_0), \mathbf{x}_\mathrm{s}(t^+)\}$. For simplicity, we assume that the control input $\mathbf{u}(t)$ remains constant over a small time interval $[t_0, t^+]$. 

Likewise, the estimated solution to the unknown end-effector dynamics~\eqref{eq:position dynamics} at any time instant $t^+ \ge t_0$ is computed as
\begin{equation}
\hat{\mathbf{x}}_\mathrm{p}(t^+) = \mathrm{ODES}\left(f_{\mathrm{p},\mathrm{NN}}, \mathbf{x}_\mathrm{p}(t_0), \mathbf{u}(t_0),[t_0,t^+]\right)
\end{equation}
where $f_{\mathrm{p},\mathrm{NN}}$ is the NN-based approximation of $f_\mathrm{p}$. To train the neural network $f_{\mathrm{p},\mathrm{NN}}$, the loss function is defined as
\begin{equation}
l_{\mathrm{p},\mathrm{NN}} = \|\hat{\mathbf{x}}_\mathrm{p}(t^+) - \mathbf{x}_\mathrm{p}(t^+)\|.
\end{equation}

\subsection{Jacobian-Based Shape-Position Hybrid Controller}

We next address the problem of hybrid shape-position control, starting with shape control. Given a reference shape state $\mathbf{x}_\mathrm{s}^\mathrm{d}(t)$, the shape control error is defined as $\mathbf{e}_\mathrm{s}(t) = \mathbf{x}_\mathrm{s}^\mathrm{d}(t) - \mathbf{x}_\mathrm{s}(t) $. We require the error dynamics to satisfy
\begin{equation}
\dot{\mathbf{e}}_\mathrm{s}(t) + \lambda_\mathrm{s} \mathbf{e}_\mathrm{s}(t) = \mathbf{0}
\label{eq:zero condition}
\end{equation}
where $\lambda_\mathrm{s} > 0$ is a constant parameter. The above condition implies that the error $\mathbf{e}_\mathrm{s}(t)$ converges to zero exponentially. Substituting the error function into~\eqref{eq:zero condition} yields
\begin{equation}
\dot{\mathbf{x}}_\mathrm{s}^\mathrm{d}(t) - \dot{\mathbf{x}}_\mathrm{s}(t) + \lambda_\mathrm{s} \left( \mathbf{x}_\mathrm{s}^\mathrm{d}(t) - \mathbf{x}_\mathrm{s}(t)\right) = \mathbf{0}.
\label{eq:expanded shape zero}
\end{equation}
To derive the controller and obtain an explicit expression for the actuation input, we also approximate the shape dynamics using a first-order Jacobian-based model:
\begin{equation}
\dot{\mathbf{x}}_\mathrm{s}(t) = \mathbf{J}_\mathrm{s}(t) \dot{\mathbf{u}}(t)
\label{eq:velocity-level shape model}
\end{equation}
where $\mathbf{J}_\mathrm{s}(t) = \partial \mathbf{x}_\mathrm{s}(t)/ \partial \mathbf{u}(t)$ is the shape-state Jacobian of the continuum robot. Substituting~\eqref{eq:velocity-level shape model} into~\eqref{eq:expanded shape zero} leads to the shape controller:
\begin{equation}
\dot{\mathbf{u}}_\mathrm{s}(t) = \mathbf{J}_\mathrm{s}^\dag(t) \left(\dot{\mathbf{x}}_\mathrm{s}^\mathrm{d}(t) + \lambda_\mathrm{s} \left( \mathbf{x}_\mathrm{s}^\mathrm{d}(t) - \mathbf{x}_\mathrm{s}(t)\right)\right)
\label{eq:shape controller}
\end{equation}
where $\mathbf{J}_\mathrm{s}^\dag(t)$ denotes the pseudo-inverse of $\mathbf{J}_\mathrm{s}(t)$.

For the end-effector position control problem, given a reference position state $\mathbf{x}_\mathrm{p}^\mathrm{d}(t)$, the position control error is defined as $\mathbf{e}_\mathrm{p}(t) = \mathbf{x}_\mathrm{p}^\mathrm{d}(t) - \mathbf{x}_\mathrm{p}(t) $. Following the above procedure for shape control, we obtain the following position controller
\begin{equation}
\dot{\mathbf{u}}_\mathrm{p}(t) = \mathbf{J}_\mathrm{p}^\dag(t) \left(\dot{\mathbf{x}}_\mathrm{p}^\mathrm{d}(t) + \lambda_\mathrm{p} \left( \mathbf{x}_\mathrm{p}^\mathrm{d}(t) - \mathbf{x}_\mathrm{p}(t)\right)\right).
\label{eq:position controller}
\end{equation}
where $\mathbf{J}_\mathrm{p}(t)$ is the position Jacobian of the robot and $\lambda_\mathrm{p} > 0$ is also a constant parameter.

For the above shape and position controllers, the reference shape and position are typically specified by the user, whereas the shape-state Jacobian and the position Jacobian remain unknown because the analytical model of the robot is unavailable. Our objective is therefore to estimate the Jacobian matrices based on the self-modeling results. Specifically, according to its definition, the estimation of the $j$-th column of the shape-state Jacobian can be expressed as
\begin{equation}
\hat{\mathbf{J}}_\mathrm{s}^{(j)}(t) = \frac{\Delta \mathbf{x}_\mathrm{s}}{2 \Delta u} = \frac{\mathbf{x}_\mathrm{s}^+ - \mathbf{x}_\mathrm{s}^-}{2 \Delta u}
\end{equation}
where $\Delta u > 0$ is a sufficiently small incremental amount, $\mathbf{x}_\mathrm{s}^+ =\mathrm{ODES}\left(f_{\mathrm{s},\mathrm{NN}}, \mathbf{x}_\mathrm{s}(t), \mathbf{u}(t) + \mathbf{E}_j \Delta u,[t,t^+]\right)$ and $\mathbf{x}_\mathrm{s}^- =\mathrm{ODES}\left(f_{\mathrm{s},\mathrm{NN}}, \mathbf{x}_\mathrm{s}(t), \mathbf{u}(t) - \mathbf{E}_j \Delta u,[t,t^+]\right)$ with $\mathbf{E}_j$ being a vector where the $j$-th element is 1 and all other elements are 0. Similarly, the estimation of the $j$-th column of the position Jacobian can be expressed as
\begin{equation}
\hat{\mathbf{J}}_\mathrm{p}^{(j)}(t) = \frac{\Delta \mathbf{x}_\mathrm{p}}{2 \Delta u} = \frac{\mathbf{x}_\mathrm{p}^+ - \mathbf{x}_\mathrm{p}^-}{2 \Delta u}
\end{equation}
where $\mathbf{x}_\mathrm{p}^+ =\mathrm{ODES}\left(f_{\mathrm{p},\mathrm{NN}}, \mathbf{x}_\mathrm{p}(t), \mathbf{u}(t) + \mathbf{E}_j \Delta u,[t,t^+]\right)$ and $\mathbf{x}_\mathrm{p}^- =\mathrm{ODES}\left(f_{\mathrm{p},\mathrm{NN}}, \mathbf{x}_\mathrm{p}(t), \mathbf{u}(t) - \mathbf{E}_j \Delta u,[t,t^+]\right)$. Therefore, the estimated shape Jacobian and position Jacobian are, respectively, obtained as\
\begin{equation}
\hat{\mathbf{J}}_\mathrm{s}(t) = \left[\hat{\mathbf{J}}_\mathrm{s}^{(1)}(t),\cdots,\hat{\mathbf{J}}_\mathrm{s}^{(m)}(t)\right]
\end{equation}
and 
\begin{equation}
\hat{\mathbf{J}}_\mathrm{p}(t) = \left[\hat{\mathbf{J}}_\mathrm{p}^{(1)}(t),\cdots,\hat{\mathbf{J}}_\mathrm{p}^{(m)}(t)\right].
\end{equation}

Finally, by combining~\eqref{eq:shape controller} and~\eqref{eq:position controller}, we propose the following shape-position hybrid controller
\begin{equation}
\dot{\mathbf{u}}(t) = \hat{\dot{\mathbf{u}}}_\mathrm{s}(t) + \hat{\dot{\mathbf{u}}}_\mathrm{p}(t) 
\label{eq:shape-position controller}
\end{equation}
where $ \hat{\dot{\mathbf{u}}}_\mathrm{s}(t)$ and $\hat{\dot{\mathbf{u}}}_\mathrm{p}(t)$ are obtained by replacing the shape-state Jacobian in~\eqref{eq:shape controller} and the position Jacobian in~\eqref{eq:position controller} with $\hat{\mathbf{J}}_\mathrm{s}(t)$ and $\hat{\mathbf{J}}_\mathrm{p}(t)$, respectively.

\subsection{Obstacle Avoidance Strategy}

As shown in Fig.~\ref{fig:method}, the position of the obstacle is denoted by $\mathbf{p}_\mathrm{obs}(t)$. Using the fitted Bézier curve, we then identify the point $\mathbf{p}_\mathrm{c}(t)$ on the curve that is closest to the obstacle. For computational convenience, the Bézier curve is uniformly discretized into $K$ points $\mathcal{O}(t)=\{\mathbf{o}_1(t),\cdots,\mathbf{o}_K(t)\}$ where the values of these points are calculated based on~\eqref{eq:multi-segment bezier}, yielding
\begin{equation}
\mathbf{p}_\mathrm{c}(t) = \mathop{\arg\min}_{\mathbf{o}_j(t) \in \mathcal{O}(t)} \|\mathbf{o}_j(t) - \mathbf{p}_\mathrm{obs}(t)\|_2.
\end{equation}

When the distance $d(t) = \|\mathbf{p}_\mathrm{c}(t) - \mathbf{p}_\mathrm{obs}(t)\|_2$ between the obstacle and the closest point falls below a predefined warning distance $d_\mathrm{w}$, the obstacle avoidance function is activated. In this case, our strategy is to apply the following escape velocity to the closest point:
\begin{equation}
\mathbf{v}_\mathrm{c}(t)=\frac{\mathbf{p}_\mathrm{c}(t) - \mathbf{p}_\mathrm{obs}(t)}{\|\mathbf{p}_\mathrm{c}(t) - \mathbf{p}_\mathrm{obs}(t)\|_2}
\end{equation}
which drives the point in the direction opposite to the obstacle until the distance $d(t)$ exceeds the warning threshold $d_\mathrm{w}$. To induce motion of the closest point along the escape direction, it is necessary to determine the required rate of change of the robot's shape state (i.e., the control points). To this end, we define a Bézier Jacobian matrix associated with the closet point:
\begin{equation}
\mathbf{J}_\mathrm{c}(t) = \frac{\partial \mathbf{x}_\mathrm{s}(t)}{\partial \mathbf{p}_\mathrm{c}(t)}
\end{equation}
which can be computed directly from the definition of the Bézier curve~\eqref{eq:multi-segment bezier}. Consequently, the escape velocity of the shape state can be obtained as follows:
\begin{equation}
\mathbf{v}_\mathrm{s}(t) = \mathbf{J}_\mathrm{c}(t) \mathbf{v}_\mathrm{c}(t)
\end{equation}
which further leads to the following escape controller:
\begin{equation}
\dot{\mathbf{u}}_\mathrm{e}(t) = \alpha(d_\mathrm{w}-d(t)) \hat{\mathbf{J}}_\mathrm{s}(t) \mathbf{v}_\mathrm{s} (t)
\end{equation}
where $\alpha > 0$ is a paremeter used to balance obstacle avoidance and shape-position control.

In the above formulation, only the escape velocity from a single view is considered. Since this work addresses three-dimensional shape control, we next propose the obstacle-avoidance controller considering escape controllers of two different views:
\begin{equation}
\dot{\mathbf{u}}_\mathrm{o}(t) = \left\{
\begin{matrix}
\dot{\mathbf{u}}^{(1)}_\mathrm{e}(t), & \text{if } d_1(t) > d_2(t) \\
\dot{\mathbf{u}}^{(2)}_\mathrm{e}(t), & \text{else}
\end{matrix}
\right.
\label{eq:obstacle controller}
\end{equation}
where $\dot{\mathbf{u}}^{(1)}_\mathrm{e}(t)$ denotes the escape controller from view 1, $d_1(t)$ is the distance between the obstacle and the closest point from view 1, $\dot{\mathbf{u}}^{(2)}_\mathrm{e}(t)$ is the escape controller from view 2, and $d_2(t)$ is the distance between the obstacle and the closest point from view 2. 

Finally, the overall controller that integrates shape-position control and obstacle avoidance is given as follows:
\begin{equation}
\dot{\mathbf{u}}(t) = \left\{
\begin{matrix}
\hat{\dot{\mathbf{u}}}_\mathrm{s}(t) + \hat{\dot{\mathbf{u}}}_\mathrm{p}(t), & \text{if } d_1(t) > d_\mathrm{w} \text{ or } d_2(t) > d_\mathrm{w} \\
\dot{\mathbf{u}}_\mathrm{o}(t) + \hat{\dot{\mathbf{u}}}_\mathrm{p}(t), & \text{else}
\end{matrix}
\right..
\label{eq:overall controller}
\end{equation}

One can also directly add the escape controllers from the two views to construct the obstacle avoidance controller. However, this strategy may slow down the execution of the primary shape-position control task, since the controller must ensure that the robot in both views moves away from the obstacle before resuming the main task. In contrast, the obstacle-avoidance controller~\eqref{eq:obstacle controller} only requires the robot in one view to escape beyond the warning distance in order to continue executing the primary task.

\begin{figure*}[tbp]
    \centering
    \includegraphics[width=1\linewidth]{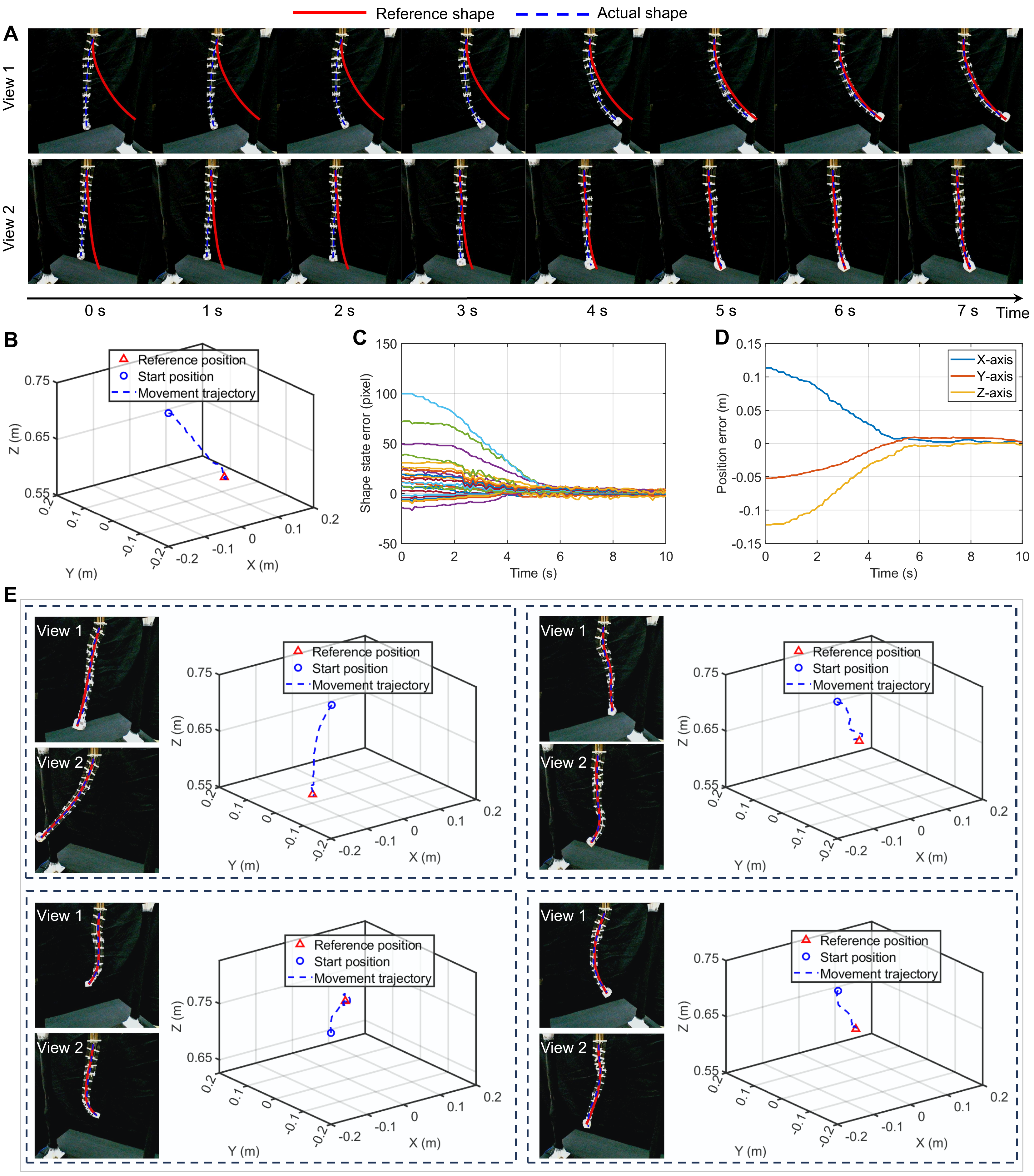}
    \caption{Experimental results of continuum robot shape-position regulation using the proposed method. (A) Experimental snapshots from two different views during the process in which the proposed method regulates the robot to the reference shape and end-effector position. (B) The robot end-effector moves from the initial position to the reference position. (C) Convergence of the shape-state regulation error to zero. (D) Convergence of the end-effector position regulation error to zero. (E) Four different shape-position regulation tasks are all successfully accomplished.}
    \label{fig:exp:regulation}
\end{figure*}

\section{Results}\label{Results}

In this section, we introduce the continuum robot platform used for method evaluation. We then assess the performance of our approach across four different tasks: shape-position regulation, shape-position tracking, shape-position regulation in the presence of obstacles, and robot self-motion, and compare it with a representative vision-based data-driven control method.

\subsection{Experimental Platform}

The experimental validation platform used in this work is illustrated in Fig.~\ref{fig:method}(F). It consists of a cable-driven, three-segment, fixed-length backbone continuum robot with a total length of 0.3 m, where the bending of each segment is actuated by four servo motors. Since the backbone length of the robot is fixed, the two opposing actuation cables within each segment are coupled, i.e., when one cable is pulled, the other is released. The proposed controller is implemented in MATLAB, and the control commands are transmitted to the motors via an Arduino Mega 2560 development board.

Two monocular cameras from different views (each costing only USD 11) are employed to capture images of the robot, with the image resolution set to $256\times 256$. For end-effector position measurement, a MicronTracker H3-60 system is used to track a marker attached to the robot tip. The robot is commanded to explore its workspace to collect 1000 data samples, which are used to train the NODE models.

\begin{figure*}[tbp]
    \centering
    \includegraphics[width=1\linewidth]{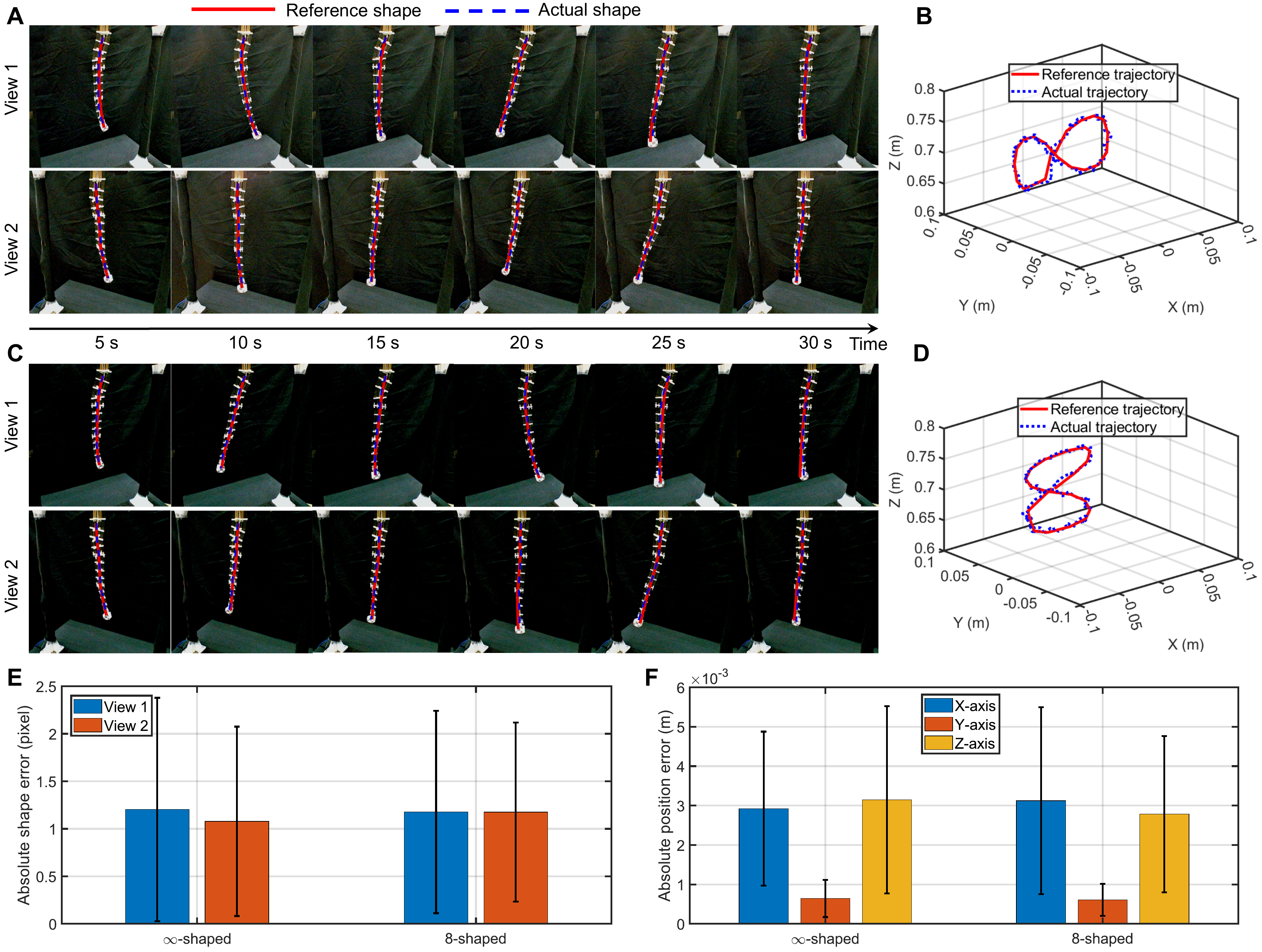}
    \caption{Experimental results of simultaneous shape-position tracking control of a continuum robot using the proposed method. (A) Experimental snapshots from two different views during tracking of the $\infty$-shaped end-effector trajectory. (B) Profile of the robot end-effector trajectory during tracking of the $\infty$-shaped trajectory. (C) Experimental snapshots from two different views during tracking of the 8-shaped end-effector trajectory. (D) Profile of the robot end-effector trajectory during tracking of the 8-shaped trajectory. (E) Distribution of shape tracking errors from different views. (F) Distribution of tracking errors along different coordinate axes.}
    \label{fig:exp:tracking}
\end{figure*}

\subsection{Task 1: Shape-Position Regulation}

We first consider the shape-position regulation task, in which the robot is driven to reach a specified shape state and end-effector position using the proposed shape-position controller~\eqref{eq:shape-position controller}. As shown in Fig.~\ref{fig:exp:regulation}(A), the red solid curves denote the reference shapes, while the blue dashed curves represent the robot's actual shapes at different time instants (i.e., the results of Bézier curve fitting). The reference shapes from two different views (View 1 and View 2) uniquely determine the robot's three-dimensional shape. It can be observed that, in both views, the actual robot shapes smoothly converge to the reference shapes.

As illustrated in Fig.~\ref{fig:exp:regulation}(B), the end effector simultaneously moves from its initial position to the reference position. Figs.~\ref{fig:exp:regulation}(C) and~\ref{fig:exp:regulation}(D) show that both the shape state error and the end-effector position error converge to nearly zero within the first 6 s. At the end of the task (i.e., at 10 s), the maximum shape error is 4 pixels (1.56\% of the image resolution), and the maximum position error is 0.003 m (1\% of the total robot length).

To further validate the proposed method, four sets of reference shapes and positions are randomly selected within the robot task space, as shown in Fig.~\ref{fig:exp:regulation}(E). In all cases, the proposed approach successfully drives the robot to the reference shapes and positions within 10 s.

In this work, the shape controller and the position controller are directly added. For scenarios where task priority must be considered (e.g., when position control has a higher priority), a hierarchical hybrid controller with task prioritization can be designed using the null space of the Jacobian matrix~\cite{Yuan2025TRO}. Such extensions are beyond the scope of this paper.

\begin{figure*}[tbp]
    \centering
    \includegraphics[width=1\linewidth]{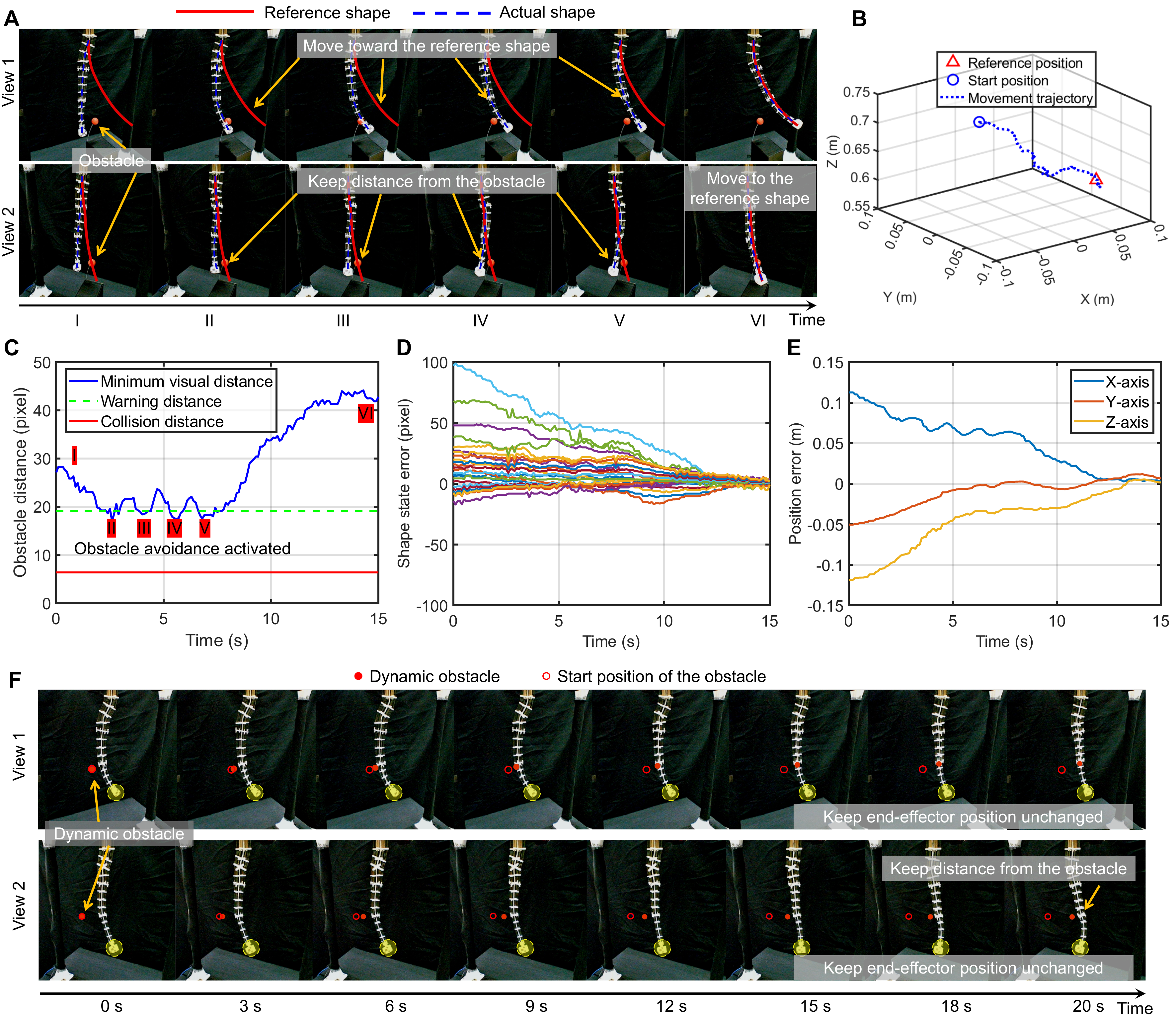}
    \caption{Experimental results of shape-position control of a continuum robot in an environment with obstacle using the proposed method. (A) Experimental snapshots from two different views during shape-position regulation in the obstacle environment. (B) Evolution of the robot end-effector position during the shape-position regulation task in the presence of obstacle. (C) Minimum visual distance between the robot body and the obstacle. (D) Shape-state regulation error during the shape-position regulation task in the obstacle environment. (E) End-effector position regulation error during the shape-position regulation task in the obstacle environment. (F) The robot avoids a dynamic obstacle by shape control while keeping the end-effector position unchanged.}
    \label{fig:exp:obstacle}
\end{figure*}

\subsection{Task 2: Shape-Position Tracking}

We next consider the shape-position tracking task, in which the robot is required to track time-varying reference shapes and end-effector positions using the proposed shape-position controller~\eqref{eq:shape-position controller}. Two sets of shape-position tracking tasks are evaluated. In the first set, the end-effector follows the $\infty$-shaped trajectory, while in the second set it follows the 8-shaped trajectory. The total duration of each task is 30 s.

As shown in Fig.~\ref{fig:exp:tracking}(A), for the $\infty$-shaped trajectory, the actual robot shapes observed from the two different views remain consistent with the reference shapes throughout the task. As illustrated in Fig.~\ref{fig:exp:tracking}(B), the actual end-effector trajectory closely follows the reference trajectory. Similar results are obtained for the 8-shaped trajectory tracking task, as shown in Figs.~\ref{fig:exp:tracking}(C) and~\ref{fig:exp:tracking}(D).

The statistical results of the shape tracking errors and end-effector position errors for both tasks are presented in Figs.~\ref{fig:exp:tracking}(E) and~\ref{fig:exp:tracking}(F). It can be observed that, across different tasks and views, the distributions of shape tracking errors are highly consistent, with no significant differences. The maximum shape error does not exceed 2.5 pixels (i.e., less than 1\% of the image resolution). For the two tasks, the distributions of end-effector position errors are also similar. However, noticeable differences can be observed among different coordinate axes. This is mainly because the $\mathrm{Y}$-axis of the coordinate frame is aligned with the vertically downward direction of the robot. Since the backbone length of the robot is fixed, its range of motion along this axis is relatively limited, resulting in smaller tracking errors compared to the other two axes. Nevertheless, the average position tracking error along the other two axes is approximately 0.003 m (1\% of the robot length), and the maximum error does not exceed 0.006 m (2\% of the robot length).

\subsection{Task 3: Obstacle-Aware Shape-Position Regulation}

The proposed framework not only enables the robot to reach the desired whole-body shape and end-effector position, but also allows additional functionalities to be realized through shape control. In this paper, obstacle avoidance is taken as an illustrative example to introduce the corresponding strategy.

We evaluate the performance of the controller~\eqref{eq:overall controller} in a shape-position regulation task under the presence of environmental obstacles. Specifically, the controller is required to regulate the robot toward the reference shape and end-effector position while exploiting shape control to avoid an obstacle. The underlying principle of the obstacle avoidance strategy is that as long as the robot body in at least one view (or both views) does not collide with the obstacle, it can be ensured that the robot does not collide with the obstacle in three-dimensional space.

Figs.~\ref{fig:exp:obstacle}(A) and~\ref{fig:exp:obstacle}(B) respectively illustrate the evolution of the robot shape and the end-effector position during task execution. Fig.~\ref{fig:exp:obstacle}(C) shows the minimum visual distance between the robot and the obstacle. It can be observed that during phase \uppercase\expandafter{\romannumeral 1}-\uppercase\expandafter{\romannumeral 2}, the visual distance remains larger than the warning distance, and the obstacle avoidance function is not activated. The robot therefore moves toward the reference shape and end-effector position. Around time instants \uppercase\expandafter{\romannumeral 2}, \uppercase\expandafter{\romannumeral 3}, \uppercase\expandafter{\romannumeral 4}, and \uppercase\expandafter{\romannumeral 5}, the minimum visual distance falls below the warning distance, triggering the obstacle avoidance mechanism. During phase \uppercase\expandafter{\romannumeral 2}-\uppercase\expandafter{\romannumeral 5}, the robot in View 1 continues to approach the reference shape normally, whereas in View 2, the obstacle avoidance controller applies an opposing escape velocity, ensuring that the robot body maintains a safe distance from the obstacle.

During phase \uppercase\expandafter{\romannumeral 5}-\uppercase\expandafter{\romannumeral 6}, since the robot body in View 1 is sufficiently far from the obstacle (i.e., beyond the warning distance), the robot in View 2 resumes moving toward the reference shape. Ultimately, the robot successfully reaches the reference shape and end-effector position while effectively avoiding the obstacle. As shown in Figs.~\ref{fig:exp:obstacle}(D) and~\ref{fig:exp:obstacle}(E), both the shape state error and the end-effector position error converge to zero.

\subsection{Task 4: Shape-Position Control for Self-Motion}

The final task is robot self-motion, in which the proposed controller~\eqref{eq:overall controller} is used to adjust the robot's shape while keeping the end-effector position fixed. As shown in Fig.~\ref{fig:exp:obstacle}(F), a dynamic obstacle approaching the robot body is considered. In both views, the red obstacle starts from its initial position and moves to the right. It can be observed that, as the obstacle approaches the robot, the obstacle avoidance function is activated, and the robot gradually adapts its shape.

Although an overlap between the obstacle and the robot is observed in View 1, the robot body in View 2 always maintains a safe distance from the obstacle, indicating that no collision occurs in three-dimensional space. More importantly, in both views, the robot end effector remains within the yellow region without noticeable deviation. These results demonstrate that the robot successfully accomplishes the self-motion task.

\begin{figure*}[tbp]
    \centering
    \includegraphics[width=0.95\linewidth]{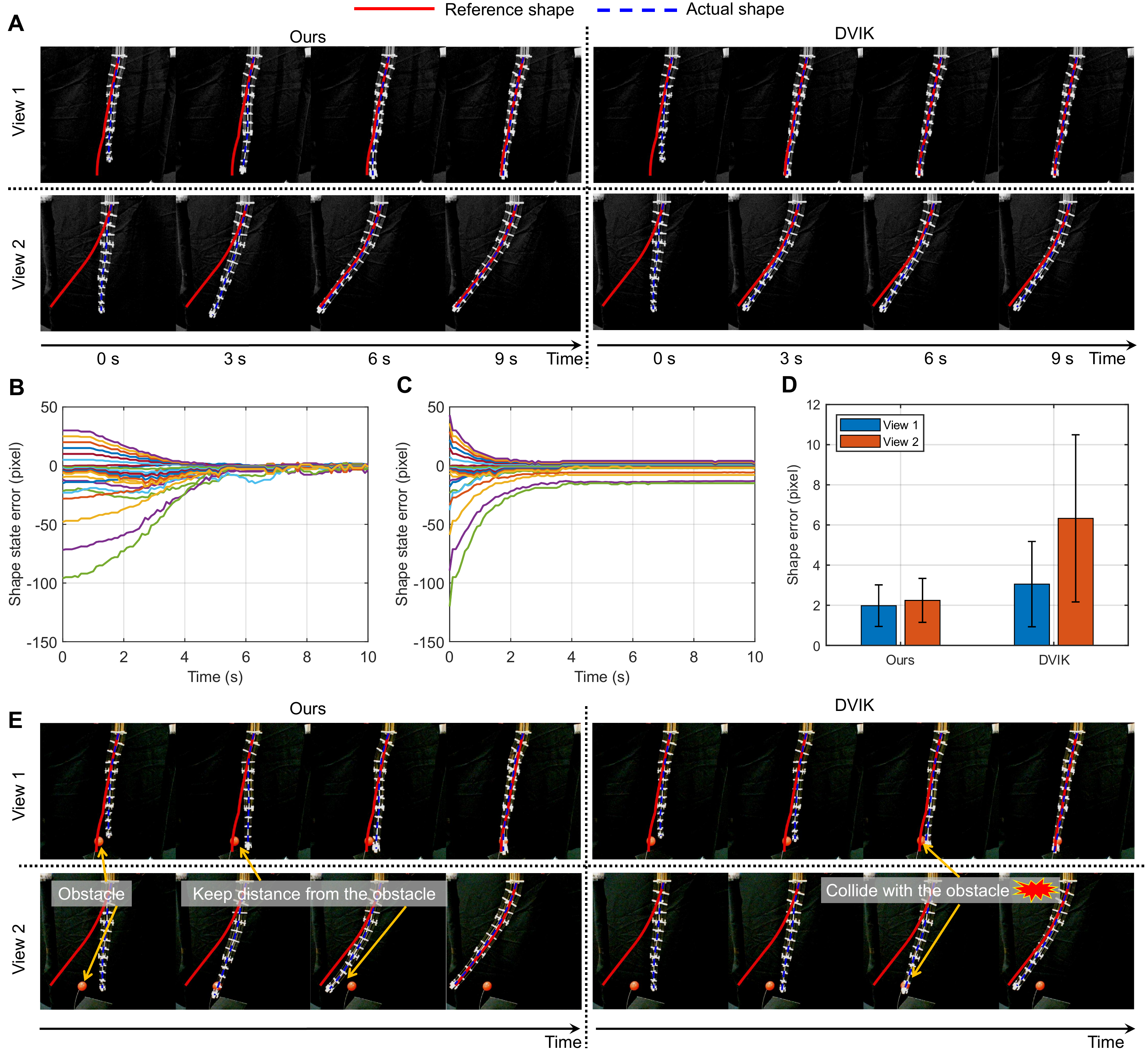}
    \caption{Comparative results of shape regulation. (A) Snapshots of the robot. Our method ensures that the robot is regulated to the reference shape in two different views, whereas the DVIK method~\cite{Almanzor2023TRO}, being a planar shape control approach, can only guarantee that the robot in View 1 reaches the reference shape. In View 2, a clear discrepancy exists between the reference shape and the actual robot shape when using the DVIK method. (B) Shape-state error of our method. (C) Shape-state error of the DVIK method. (D) Statistical error results for 12 different shape regulation tasks. The proposed method achieves nearly identical error distributions in the two views, and these errors are consistently smaller than those of the DVIK method in View 1. Moreover, since the DVIK method performs planar shape regulation, its error in View 2 is significantly larger than that in View 1. (E) Shape regulation results in the environment with an obstacle. The proposed method can simultaneously exploit interpretable shape state information during shape regulation to improve the controller and achieve obstacle avoidance, whereas the DVIK method fails to avoid the obstacle, resulting in collisions between the robot and the obstacle.}
    \label{fig:exp:comparison}
\end{figure*}

\subsection{Comparison}

Finally, we quantitatively compare the proposed method with the deep visual inverse kinematic (DVIK) approach proposed by Almanzor et al.~\cite{Almanzor2023TRO}, as both methods are data-driven and based on visual images. Since the DVIK method only addresses shape control, we compare it with the shape controller proposed in this work. Both methods are used to drive the robot toward the same reference shape.

Because the DVIK method performs planar two-dimensional shape control, only the images from View 1 are used for control, while the results from View 2 are used solely for comparative analysis. As shown in Fig.~\ref{fig:exp:comparison}(A), the proposed method enables the robot shapes observed from both views to converge to the reference shape. Although the DVIK method can drive the robot to the reference shape in View 1, it is inherently a single-view control approach. As a result, apparent discrepancy between the actual and reference shapes can be observed in View 2. In the worst case, single-view control may lead to substantial errors in the other view due to the non-unique mapping between planar observations and the three-dimensional robot shape.

Figs.~\ref{fig:exp:comparison}(B) and~\ref{fig:exp:comparison}(C) illustrate the shape state errors for the two methods, showing that the proposed method achieves a significantly smaller steady-state error than the DVIK approach. To further ensure the reliability of the comparison, we randomly select 12 different reference shape states to evaluate both methods, and the statistical results are summarized in Fig.~\ref{fig:exp:comparison}(D). It can be seen that the proposed method yields similar error distributions across the two views. In contrast, the DVIK method exhibits larger errors in View 1 compared to the proposed method, and, more importantly, the errors observed in View 2 are significantly higher than those in View 1.

We also evaluate the performance of both methods in the presence of an environmental obstacle, as shown in Fig.~\ref{fig:exp:comparison}(E). As expected, the proposed method is able to regulate the robot to the reference shape while simultaneously avoiding the obstacle in the environment. In contrast, the DVIK method causes the robot to collide with the obstacle during the shape regulation process, since it is designed for obstacle-free environments and cannot exploit shape control to actively perform obstacle avoidance.

The reasons for selecting the DVIK method for quantitative comparison are as follows.
First, model-based approaches require explicit physical knowledge to construct analytical models, whereas the proposed method does not rely on any prior assumptions or physical modeling. Second, shape control methods based on discrete point representations typically involve substantially different hardware setups and sensing requirements compared to our approach. These methods often require a large number of markers or sensors distributed along the robot body to enable shape control, while our method achieves shape control purely from visual observations without such instrumentation. As a result, a strict and fair comparison with these approaches is not feasible. In contrast, the DVIK method closely matches our experimental setup. It is a data-driven approach that does not depend on physical models or body-mounted markers and relies solely on visual images of the robot to achieve control. Although a recent general controller is also capable of image-based shape control~\cite{Tang2026SA}, it primarily focuses on improving adaptability across different tasks and perturbations. Similar to DVIK, the general controller is also based on a single-view visual input and does not provide an interpretable representation of the robot shape. By contrast, the proposed method explicitly emphasizes shape interpretability, which enables the effective exploitation of shape control for more complex behaviors, such as obstacle avoidance and self-motion.

\section{Discussion and Conclusion}\label{Discussion}

This paper proposes a novel shape-interpretable visual self-modeling  framework for continuum robots, enabling hybrid shape-position control without requiring analytical kinematic or dynamic models. By encoding robot shapes from multi-view planar images with Bézier-based shape parameterization and learning both shape and position dynamics using neural ordinary differential equations, the proposed approach achieves unified modeling, geometry-aware control, and obstacle avoidance within a single data-driven framework. The effectiveness of the proposed framework is validated experimentally on a cable-driven, three-segment continuum robot. The results demonstrate accurate shape-position regulation and tracking, while successfully avoiding obstacles.

The proposed approach offers several distinct advantages when compared with existing methods for continuum robot modeling and control. First, compared with model-based approaches, the proposed framework does not rely on analytical kinematic or dynamic models. Continuum robots are characterized by strong nonlinearity, high redundancy, and significant parameter uncertainty, which make accurate model derivation and real-time computation extremely challenging. By leveraging vision-based self-modeling and data-driven learning, the proposed method avoids these modeling burdens while still achieving accurate shape-position control, thereby improving practical applicability. Second, compared with data-driven shape control methods based on discrete point measurements, such as those relying on multiple markers attached to the robot body, the proposed approach eliminates the dependence on dense physical sensing or extensive labeling. Instead of tracking a large number of discrete points, robot shapes are encoded using Bézier curves derived directly from visual observations, resulting in a compact yet expressive shape representation. This significantly reduces system complexity, sensing requirements, and setup effort, while preserving sufficient geometric information for shape control. Third, compared with end-to-end vision-based data-driven control methods, the proposed approach provides several fundamental advantages. End-to-end approaches typically employ implicit latent representations, which lack explicit geometric semantics and limit their ability to reason about robot-environment interactions. In contrast, the proposed method achieves (\romannumeral1) an increased shape dimensionality (from planar space to 3D space) through multi-view shape encoding, (\romannumeral2) an explicit, geometry-aware and control-oriented shape representation that enables direct perception of robot-environment distances, and (\romannumeral3) a unified hybrid shape-position control framework that allows shape regulation to be exploited for advanced behaviors such as obstacle avoidance and self-motion. These properties make the proposed method more flexible, interpretable, and suitable for safety-critical tasks.

The superior performance of the proposed method over the baseline method stems from its explicit, interpretable shape representation embedded in the data-driven control framework. Unlike end-to-end vision-based controllers, where robot shape and environment geometry remain latent, the proposed approach reconstructs the robot shape using low-dimensional, continuous curve parameters, making the geometric relationship between the robot and surrounding obstacles directly observable. This explicit observability enables obstacle avoidance and self-motion to be naturally integrated into the control law. Moreover, by jointly exploiting multiple views, the proposed method resolves the inherent ambiguity of single-view shape control and ensures consistency across different views. In contrast, existing methods often suffer from non-unique visual-to-shape mappings or rely on discrete, marker-based representations that lack continuity and scalability.

The proposed vision-based, shape-interpretable control framework suggests a new paradigm for continuum robot control that bridges data-driven learning and geometry-aware modeling. By explicitly representing robot shape with low-dimensional, physically meaningful parameters extracted from visual observations, the proposed method enables environment-aware control behaviors that go beyond conventional goal-reaching tasks, such as obstacle avoidance and self-motion. This elevates robot shape from a latent variable in end-to-end learning to a task-relevant and perceptually grounded control primitive. As a result, the framework provides a general and extensible foundation for safe and flexible interaction between continuum robots and complex environments, and can potentially be extended to other soft or highly deformable robotic systems.

Despite its strengths, the proposed framework has several limitations that suggest directions for future research. First, the current approach relies on accurate planar image segmentation and skeleton extraction, which may be sensitive to lighting, severe occlusion, or background complexity. Integrating advanced or adaptive vision segmentation techniques could improve robustness in unstructured environments. Second, the Bézier-curve-based shape representation, while compact and control-oriented, imposes smoothness and may not fully capture localized high-curvature deformations. Future work could explore adaptive or piecewise parametric representations to better describe complex shapes. Third, although multi-view planar images are used to uniquely determine 3D shape, the framework does not explicitly reconstruct the full three-dimensional configuration, limiting applications that require precise 3D perception. Extending the method to include explicit 3D reconstruction from visual cues is a promising direction. Finally, the current shape-position control scheme treats shape and position tasks with equal priority, which may not be optimal for scenarios with hierarchical task requirements. Incorporating task-priority frameworks, such as null-space projection or weighted optimization, could enable more flexible and safe task execution. Addressing these limitations will further enhance the autonomy, adaptability, and applicability of continuum robots in complex, dynamic environments.

Overall, this work presents a vision-based, shape-interpretable framework for continuum robot modeling and control. This framework provides a foundation for more autonomous, adaptable, and safe continuum robot systems, and suggests a generalizable paradigm for integrating visual perception with geometry-aware control in soft and highly deformable robotics.

\bibliographystyle{ieeetr}
\bibliography{reference}

\end{document}